\documentclass[runningheads]{llncs}
\usepackage[T1]{fontenc}
\usepackage{graphicx}
\usepackage{hyperref}
\hypersetup{
colorlinks   = true,
citecolor    = blue,
urlcolor    =  blue
}
\usepackage{color}

\usepackage{amsmath,amssymb,amsfonts}
\usepackage{algorithmic}
\usepackage{algorithm2e}
\usepackage{graphicx}
\usepackage{textcomp}
\usepackage{xcolor}
\usepackage{listings}
\usepackage{comment}
\usepackage{caption}
\usepackage{subcaption}
\newcommand{\footURL}[1]{\footnote{\url{#1}}}
\usepackage[utf8]{inputenc}
\usepackage{threeparttable}
\usepackage{multirow}
\usepackage{hhline}
\usepackage{makecell}
\usepackage{pdflscape}
\usepackage{afterpage}
\usepackage{capt-of}
\usepackage{adjustbox}

\usepackage{color}
\usepackage{colortbl}
\usepackage[numbers,sort&compress]{natbib}

\usepackage{tabularx}
\newcolumntype{Y}{>{\centering\arraybackslash}X}
\newcolumntype{Z}{>{\raggedleft\arraybackslash}X}

\def\BibTeX{{\rm B\kern-.05em{\sc i\kern-.025em b}\kern-.08em
    T\kern-.1667em\lower.7ex\hbox{E}\kern-.125emX}}

\newcommand{\Si}{\texttt{Si}}
\newcommand{\En}{\texttt{En}}
\newcommand{\Ta}{\texttt{Ta}}
\newcommand{\Es}{\texttt{Es}}
\newcommand{\Fr}{\texttt{Fr}}
\newcommand{\De}{\texttt{De}}
\newcommand{\It}{\texttt{It}}
\newcommand{\Zh}{\texttt{Zh}}
\newcommand{\Ja}{\texttt{Ja}}
\newcommand{\Tr}{\texttt{Tr}}
\newcommand{\EnSi}{\texttt{En-Si}}

\begin{document}
\title{How Good is BLI as an Alignment Measure: A Study in Word Embedding
Paradigm}
\titlerunning{How Good is BLI as an Alignment Measure}
%
\author{Kasun Wickramasinghe\orcidID{0009-0003-5972-8012} \and
Nisansa de Silva\orcidID{0000-0002-5361-4810}}
%
\authorrunning{K. Wickramasinghe and N. de Silva}
%
\institute{Dept. of Computer Science \& Engineering, University of Moratuwa, Sri Lanka \\
\email{\{kasunw.22,NisansaDdS\}@cse.mrt.ac.lk}}
\maketitle              
\begin{abstract}
Sans a dwindling number of monolingual embedding studies originating predominantly from the low-resource domains, it is evident that multilingual embedding has become the de facto choice due to its adaptability to the usage of code-mixed languages, granting the ability to process multilingual documents in a language-agnostic manner, as well as removing the difficult task of aligning monolingual embeddings. But is this victory complete? Are the multilingual models better than aligned monolingual models in every aspect? Can the higher computational cost of multilingual models always be justified? Or is there a compromise between the two extremes? 
Bilingual Lexicon Induction (BLI) is one of the most widely used metrics in terms of evaluating the degree of alignment between two embedding spaces.
In this study, we explore the strengths and limitations of BLI as a measure to evaluate the degree of alignment of two embedding spaces. 
Further, we evaluate how well traditional embedding alignment techniques, novel multilingual models, and combined alignment techniques perform BLI tasks in the contexts of both high-resource and low-resource languages. In addition to that, we investigate the impact of the language families to which the pairs of languages belong. 
We identify that BLI does not measure the true degree of alignment in some cases and we propose solutions for them. We propose a novel stem-based BLI approach to evaluate two aligned embedding spaces that take into account the inflected nature of languages as opposed to the prevalent word-based BLI techniques. Further, we introduce a vocabulary pruning technique that is more informative in showing the degree of the alignment, especially performing BLI on multilingual embedding models.
In addition to that, we find that in most cases, combined embedding alignment techniques that make use of both traditional alignment and multilingual embeddings perform better while for certain scenarios multilingual embeddings perform better (especially low-resource language cases). The traditional aligned embeddings lag behind the other two types of aligned embeddings in the majority of the cases.

\end{abstract}
\section{Introduction} \label{introduction}
Numerical representation is needed to perceive anything to a computer. When it comes to images pixel-based representation is used and, in contrast in the language case, vector representations such as one-hot-encoding, TF-IDF~\cite{salton1988term} or embedding representations such as Word2Vec~\cite{mikolov2013efficient}, GloVe~\cite{pennington-etal-2014-glove}, FastText~\cite{bojanowski-etal-2017-enriching}, ELMo~\cite{peters-etal-2018-deep}, BERT~\cite{devlin-etal-2019-bert} are used. The main difference between the vector and embedding representations is that the vector representations are handcrafted while the embedding representations are auto-learned vector representations by optimizing one or more particular criteria. 
The contributions of this study are,
\begin{itemize}
    \item Study how good BLI is as a Measure of the degree of Alignment
    \item Introduce a stem-based BLI technique for inflected language evaluation
    \item Introduce a vocabulary pruning technique for multilingual embedding-based BLI
    \item Study multilingual embedding models in word embeddings paradigm
    \item Compare aligned monolingual models, multilingual models, and, a combination of these two models in bilingual lexicon induction (BLI) tasks
    \item Study the behavior of them in high-resource and low-resource language cases~\cite{ranathunga-de-silva-2022-languages}
    \item Study the impact of aligning languages of two language families
    \item Release the code base and findings in GitHub\footURL{https://github.com/kasunw22/sinhala-word-embedding-alignment}
\end{itemize}

As far as the multilingual or cross-lingual tasks are concerned, the embeddings of different languages should be aligned or, in other words, the embeddings of different languages should share a common embedding space~\cite{mikolov2013exploiting, joulin-etal-2018-loss}. 
The monolingual models are bound to a particular language. Each language should have a separately trained embedding model which makes it difficult to perform multilingual embedding arithmetics since the embeddings are not aligned due to independently trained models.
We can adopt monolingual embedding models for cross-lingual tasks by aligning different monolingual models~\cite{mikolov2013exploiting, xing-etal-2015-normalized, smith2016offline, joulin-etal-2018-loss}. Multilingual models by default share a common embedding space for all the languages that the model supports. This is achieved by mainly two techniques. One is having a multilingual vocabulary and encouraging embedding alignment through the training process using special learning objectives such as Multilingual Masked Language Modelling (mMLM)~\cite{conneau2019cross, conneau-etal-2020-unsupervised} and Translation Language Modeling (TLM)~\cite{feng-etal-2022-language, conneau2019cross}. Depending on the training objective, multilingual models may use multilingual data (mMLM) or parallel multilingual data (TLM). The other method is training a student model to produce aligned multilingual embeddings using multiple monolingual teacher models through knowledge distillation~\cite{reimers-gurevych-2020-making, artetxe-schwenk-2019-massively, heffernan-etal-2022-bitext}.

It is not intuitive to get a direct measurement for the degree of alignment between two or more embedding spaces. Instead what is commonly used are indirect measurements which are very common in scientific studies. The common such measurements are bilingual lexicon induction (BLI) or word translation task~\cite{irvine-callison-burch-2017-comprehensive}, cross-lingual natural language inference (XNLI)~\cite{conneau-etal-2018-xnli}, cross-lingual semantic word similarity~\cite{conneau2017word}, sentence translation retrieval~\cite{conneau2017word,velayuthan2025encoder} and cross-lingual question-answering~\cite{conneau-etal-2020-unsupervised}. Among those different measures, BLI is the most common measure as far as aligned word embeddings are concerned. BLI is widely used but the limitations and edge conditions of BLI as an alignment measure are not well studied. Our main objective of this paper is to explore that.



\section{Related Work}



\subsection{Multilingual Embeddings}

Multi-lingual embedding models generate embeddings irrespective of a language~\cite{ahuja-etal-2022-calibration, cooper-stickland-etal-2023-robustification}. \emph{mBERT}~\cite{devlin-etal-2019-bert, reimers-gurevych-2020-making}, \emph{XLM}~\cite{conneau2019cross}, \emph{XLM-R}~\cite{conneau-etal-2020-unsupervised} \emph{LASER}~\cite{artetxe-schwenk-2019-massively, heffernan-etal-2022-bitext} and \emph{LaBSE}~\cite{feng-etal-2022-language} are popular multilingual models.
We find two main methods of training multilingual models. Certain models have used multiple monolingual models to extend knowledge for building a multilingual model through knowledge distillation (mBERT~\cite{reimers-gurevych-2020-making}, LASER~\cite{artetxe-schwenk-2019-massively, heffernan-etal-2022-bitext}) while some other models have used large corpora of monolingual and multilingual parallel datasets to pre-train large multilingual models using training objectives such as Multilingual Masked Language Modelling (mMLM) and Translation Language Modeling (TLM) (LaBSE~\cite{feng-etal-2022-language}, XLM~\cite{conneau2019cross}, XLM-R~\cite{conneau-etal-2020-unsupervised})

\subsection{Word Embedding Alignment} \label{embedding_alignment_techniques}


The simplest method of embedding alignment is to find a linear mapping between two unaligned embedding spaces~\cite{mikolov2013exploiting} by simply minimizing the $l2$ norm of the error.
A slightly improved version of this method is finding an orthogonal mapping assuming normalized constraint on the embeddings~\cite{xing-etal-2015-normalized}.
Another method is called \emph{Orthogonal Procrustes Mapping} where the mapping is found using the singular value decomposition (SVD) of the product of the two normalized embedding spaces~\cite{smith2016offline, conneau2017word}.
One of the better techniques out there is optimizing the \emph{Cross-domain similarity local scaling} (CSLS) loss as the optimization criterion~\cite{joulin-etal-2018-loss} which tries to address the so-called \emph{hubness problem} in embedding alignment by ensuring a symmetric mapping in the objective.
\textit{VecMap}\footURL{https://github.com/artetxem/vecmap} is another promising technique that proposes a series of linear transformations to align two embedding spaces~\cite{artetxe2018aaai}. VecMap formalizes the linear mapping.
\citet{li-etal-2022-improving} have used the \textit{InfoNCEloss}~\cite{oord2018representation} to iteratively improve the aligned word embeddings. They propose a two-stage contrastive learning-based pipeline. Stage-1 is based on the VecMap~\cite{artetxe2018aaai} alignment procedure with a self-learning stage by optimizing the InfoNCEloss.
Stage 2 is a contrastive fine-tuning for mBERT embeddings with the help of stage-1 aligned embeddings and finally, the stage-1 aligned embeddings are again mapped to the mBERT aligned space. The final word embedding for a given word is calculated using a linear combination of re-mapped stage-1 and stage-2 embeddings. These two stages are referred to as C1 and C2 in the paper. C2 has shown state-of-the-art results in standard BLI tasks. 

All the above methods are supervised but, there are some promising unsupervised techniques such as unsupervised VecMap~\cite{artetxe-etal-2018-robust}, 
Wasserstein Procrustes Analysis~\cite{grave2019unsupervised} and Quantized Wasserstein Procrustes Analysis~\cite{aboagye-etal-2022-quantized}.
There are some refinement techniques as well for further tuning the alignment after the initial mapping~\cite{artetxe-etal-2017-learning, conneau2017word, doval-etal-2018-improving}.


\subsection{Aligned Global Embeddings vs Multilingual Contextualized Embeddings}
\citet{saadi-etal-2022-comparative} have conducted a comprehensive analysis of using the contextualized embeddings for cross-lingual tasks as a substitution for aligned global word embeddings. They have experimented with the quality of aligned contextualized embeddings for bilingual-lexicon induction (BLI)~\cite{irvine-callison-burch-2017-comprehensive}, word retrieval, and cross-lingual natural language inference (XNLI)~\cite{conneau-etal-2018-xnli} tasks. 

In another study, \citet{zhang-etal-2021-combining} have experimented with a method to align global and contextualized monolingual word embeddings to utilize the advantages of both embedding types.
Work by \citet{li-etal-2022-improving} and \citet{hammerl-etal-2022-combining} have combined both global and contextualized embeddings for better-aligned embeddings. As mentioned in Section~\ref{embedding_alignment_techniques}, \citet{li-etal-2022-improving} have improved traditionally aligned embeddings further by mapping them again with contextualized embeddings. 
\citet{hammerl-etal-2022-combining} claim to extract \emph{partially aligned static embeddings} from XLM-R and then further align them using VecMap. They then train a contextualized model to generate more aligned multilingual embeddings.

\subsection{Cross-Lingual Embedding Evaluation Techniques}
There are many tasks used by the research community to evaluate the quality of cross-lingual embeddings. One of the most popular methods is bilingual lexicon induction (BLI) or the word translation task~\cite{irvine-callison-burch-2017-comprehensive}. Among the other tasks for this purpose are; cross-lingual natural language inference (XNLI)~\cite{conneau-etal-2018-xnli}, cross-lingual semantic word similarity~\cite{conneau2017word}, sentence translation retrieval~\cite{conneau2017word}, and cross-lingual question-answering~\cite{conneau-etal-2020-unsupervised}.

\subsection{BLI Benchmarking Datasets}
MUSE\footURL{https://github.com/facebookresearch/MUSE}~\cite{conneau2017word} is one of the largest bilingual dictionary collections with 110 language pairs.However the usability of this data set is limited by the fact that 90 of the language pairs consist of English as one of the members and thus there are only 20 non-English language pairs.
XLing\footURL{https://github.com/codogogo/xling-eval}~\cite{glavas-etal-2019-properly} is another BLI dataset consisting of 8 languages and 56 BLI directions in total.
Comparatively, PanLex-BLI~\footURL{https://github.com/cambridgeltl/panlex-bli}~\cite{vulic-etal-2019-really} is a large-scale BLI dataset that consists of 210 BLI directions of 15 low-resource languages.

\section{Methodology}

We are evaluating our objectives using a cross-lingual evaluation task based on the model type, inflection of the language, mixed-code vocabularies, and language family. We have carefully selected 10 languages covering all these cases. 
For alignment evaluation, we use FastText embeddings for traditional embedding alignment and FastText vocabularies as vocabularies for all the languages we evaluate. 

\subsection{BLI to Measure the Degree of Alignment}
Our aim in this study is to check how well multilingual embeddings are aligned. To quantitatively evaluate this, one of the commonly used tasks by the monolingual word embedding alignment researchers is finding the~\emph{word translation precision} or \emph{Bilingual Lexicon Induction (BLI)} score (we have used these two terms interchangeably in the paper), which checks from a parallel test set, how many target translations of source words can be found using the aligned embedding spaces~\cite{conneau2017word, joulin-etal-2018-loss, smith2016offline, wickramasinghe-de-silva-2023-sinhala, irvine-callison-burch-2017-comprehensive}. 
BLI is one of the most common yet powerful methods used to compare how aligned two languages are~\cite{conneau2017word, joulin-etal-2018-loss, smith2016offline, severini-etal-2022-towards, saadi-etal-2022-comparative, li-etal-2022-improving, artetxe2018aaai}. Due to its wild spread use, its positives, negatives, and limitations are well-studied in the literature~\cite{sogaard-etal-2018-limitations, zhang-etal-2021-combining, vulic-etal-2019-really}.
We have used \emph{MUSE}~\cite{conneau2017word}, which is a free and open-source dataset for all the language pairs we evaluated other than the \EnSi{} pair since that pair is not available in MUSE or any other major BLI benchmarking datasets. For the \EnSi{} pair, we used another open-source dataset provided by \citet{wickramasinghe-de-silva-2023-sinhala}, which has been created considering the same criteria as MUSE. The reason for selecting $\Si$ as one of the languages in our study is that it is considered to be a low-resource inflected language~\cite{de2019survey}.

\subsection{Multilingual Models for Aligned Word Embeddings}

Multilingual models are becoming more common~\cite{ahuja-etal-2022-calibration, cooper-stickland-etal-2023-robustification} for cross-lingual tasks compared to using aligned monolingual models irrespective of the factor of computational cost~\cite{kim2021scalable, tang-etal-2021-multilingual}. The main advantage of using multilingual models for cross-lingual tasks is that all the languages supported by the model are already aligned~\cite{feng-etal-2022-language, conneau2019cross,conneau-etal-2020-unsupervised} and no overhead is applied for explicit alignment of different languages. In this work, we have experimented with how good the word embeddings of multilingual models are compared to the aligned traditional word embeddings for a number of languages.

\subsection{Aligned Monolingual Models vs Multilingual Models} \label{mono_vs_multi}
Multilingual models are language-agnostic, given that in them, all the supported languages share a common embedding space as we discussed in Section~\ref{introduction}. In multilingual model training, due to the significant imbalance of training data for different languages, different techniques have been followed to reduce the bias towards high-resource languages and to balance the quality between high-resource and low-resource languages~\cite{conneau2019cross, feng-etal-2022-language}. In the pursuit of this goal, the accuracy obtained for certain languages can be compromised in favour of balance. Even at that cost, the data imbalance effect cannot be completely negated by the training tricks that are followed~\cite{wickramasinghe-de-silva-2023-sinhala}. Under this scenario, in this study, we are investigating the cases where aligned-monolingual models are more beneficial than multilingual models and vice-versa.

\subsection{Effect of Inflection} \label{inflected_lang_evaluation}

We have used English (\En) as the reference language in our experiments so that all the results can be compared on a common basis. 
An inflected language, also known as a fusional language, is a language that uses modifications to words themselves to indicate grammatical information. This contrasts analytic languages, which rely more on word order to show grammatical relationships~\cite{crystal2011dictionary}. Most languages have a certain degree of inflection; but certain languages, such as Sinhala (\Si), are considered to be highly inflected~\cite{de2019survey,wickramasinghe2023sinhala1,de-mel-etal-2025-sinhala} and therefore contain an above-average number of variations for a given single word. 

For inflected languages, even when the model alignment happens properly, it may not be properly reflected through the BLI task due to not having exact matches as expected in the test sets. Therefore, we experimented with a soft matching rather than performing an exact match. We refer to this as \emph{Stem-based BLI}. 
Here, in addition to finding the exact target translation for a given source word from the test set, we query for the \emph{stem} of the expected target translation word \emph{if an exact match is not found}.
This is achieved as shown in Algorithm~\ref{stem_based_seach_algo}. 

\begin{algorithm}[!htb]
\caption{Algorithm to perform stem-based BLI}
\KwData{$dataset \gets$ BLI evaluation dataset}
$count_{total} \gets 0$\;
$count_{correct} \gets 0$\;
\While{$datapoint \neq None$}{
  $word_{src}, word_{tgt} = datapoint.split()$\;
  $tgt_{topN} \gets GetTopN(word_{src}, N)$\;
  \eIf{$word_{tgt}$ in $tgt_{topN}$}{
    $count_{correct} \gets count_{correct} + 1$\;
  }{
  $word_{tgt-stem} \gets stem(word_{tgt})$\;
  $tgt_{topN-stem} \gets stem(tgt_{topN})$\;
  $tgt_{full} \gets tgt_{topN} \cup tgt_{topN-stem}$\;
  \If{$word_{tgt}$ in $tgt_{full}$ || $word_{tgt-stem}$ in $tgt_{full}$}{
      $count_{correct} \gets count_{correct} + 1$\;
    }
  }
  $count_{total} \gets count_{total} + 1$\;
  $datapoint \gets next(dataset)$\;
}
$score_{BLI} \gets count_{correct} / count_{total}$\;
\label{stem_based_seach_algo}
\end{algorithm}

\subsection{Pruning Vocabularies} \label{pruning_method}
The vocabulary of most embedding models is not pure. They also contain words/tokens of some other languages due to human usage of different languages together (\textit{code-mixed}). When it comes to BLI, code-mixed usage can negatively affect the results if we consider only the top-1 match. This issue is further exacerbated in multilingual models since multilingual models could give the top-1 match from any language. Thus, we wanted to evaluate how good the alignment is considering only the eponymous language of the model. 
The simple solution we propose for this issue is pruning the vocabulary by removing tokens that do not belong to the language of interest prior to the BLI evaluation. 
The pruning can be conducted in a straightforward manner for languages that use non-Latin scripts. Such is the case for Russian (\texttt{Ru}), Sinhala (\Si), Chinese (\Zh), Tamil (\Ta), and Japanese (\Ja) where we simply eliminate the characters in the ASCII range and conduct the evaluation.

\subsection{Effect of the Language Family}
We have chosen the languages in this study to also enable us to observe the effect of the relatedness of the languages. We have chosen English (\En), Spanish (\Es), French (\Fr), German (\De), Italian (\It), Russian (\texttt{Ru}), and Sinhala (\Si) from the \textit{Indo-European} family to which most of the human written languages belong. Given the vastness of this language tree, we have strived to pick languages representing a number of its most populous sub-trees. As such, English (\En) and German (\De) represent the \textit{Germanic} sub-tree, Spanish (\Es), French (\Fr), and Italian (\It) represent the \textit{Romance} sub-tree, Russian (\texttt{Ru}) represents the \textit{Slavic} sub-tree, and Sinhala (\Si) represents the \textit{Indic} sub-tree. As for the languages that do not belong to the \textit{Indo-European} family tree; Chinese (\Zh) belongs to the \textit{Sino-Tibetan} family, Tamil (\Ta) belongs to the \textit{Dravidian} family, Japanese (\Ja) belongs to the \textit{Japonic} family, and finally the Turkish (\Tr) belongs to the \textit{Turkic} family. 

\section{Experiments}\label{experiments}

The training sets (datasets used for supervised embedding alignment of FastText embeddings) consist of 5000 unique source words and the test sets (datasets used to evaluate aligned embeddings using the BLI task) contain 1500 unique source words. 
Four multilingual embedding models were first shortlisted for experimentation based on the state-of-the-art accuracy they have claimed in their publications~\cite{reimers-gurevych-2020-making, conneau-etal-2020-unsupervised, artetxe-schwenk-2019-massively, heffernan-etal-2022-bitext, feng-etal-2022-language} and the extended support for many languages.

\begin{enumerate}
    \item mBERT\footURL{https://bit.ly/3NtVoRv}~\cite{devlin-etal-2019-bert, reimers-gurevych-2020-making}
    \item XLM-R\footURL{https://bit.ly/3v2lO6k}~\cite{conneau-etal-2020-unsupervised}
    \item LASER~\footURL{https://github.com/facebookresearch/LASER} 2 and 3~\cite{artetxe-schwenk-2019-massively, heffernan-etal-2022-bitext}
    \item LaBSE\footURL{https://bit.ly/471KcSZ}~\cite{feng-etal-2022-language}
\end{enumerate}

For evaluation shown in Table~\ref{different_multilingual_models_comparison}, we experimented with three languages belonging to three different language families. (\Es$\rightarrow$Indo-European, \Zh$\rightarrow$Sino-Tibetan, \Tr$\rightarrow$Turkic). According to the results in Table~\ref{different_multilingual_models_comparison}, we selected best out of these four, \textit{LaBSE} as the candidate multilingual model for our subsequent experiments with traditional alignment techniques. 
%

\begin{table*}[!ht]
    \caption{Word translation precision of different multilingual models. For Zh$\rightarrow$En, Zh vocabulary has been \emph{pruned} as per Section~\ref{pruning_method}}
    \label{different_multilingual_models_comparison}
    \centering
    \resizebox{1\columnwidth}{!}{%
    \begin{tabular}{l|l|ccc|ccc|ccc|ccc}
    \hline
        \multirow{3}{*}{\makecell{Lang\\\texttt{X}}} & \multirow{3}{*}{Method} & \multicolumn{6}{c|}{\En\texttt{-X}} & \multicolumn{6}{c}{\texttt{X-}\En} \\
        \hhline{~~------------}
         & & \multicolumn{3}{c|}{NN} & \multicolumn{3}{c|}{CSLS} & \multicolumn{3}{c|}{NN} & \multicolumn{3}{c}{CSLS} \\
        \hhline{~~------------}
         & & P@1 & P@5 & P@10 & P@1 & P@5 & P@10 & P@1 & P@5 & P@10 & P@1 & P@5 & P@10 \\  \hline

        \multirow{3}{*}{\Es} & mBERT & 36.1 & 74.4 & 82.0 & 25.7 & 58.7 & 70.9 & 50.4 & 72.5 & 79.7 & 49.4 & 76.2 & 82.1 \\

        & XLM-R & 37.4 & 78.2 & 83.7 & 28.2 & 62.7 & 73.1 & 50.9 & 71.4 & 76.5 &  61.3 & 78.1 & 80.9 \\ 
        & LASER2/3 & 37.9 & 75.3 & 82.1 & 39.3 & 75.3 & 82.3 & 57.6 & 78.9 & 85.1 & 56.2 & 78.2 & 84.6 \\
        & LaBSE & \textbf{40.0} & \textbf{83.1} & \textbf{89.6} & \textbf{40.5} & \textbf{84.7} & \textbf{90.2} & \textbf{65.3} & \textbf{88.9} & \textbf{92.1} & \textbf{65.0} & \textbf{87.7} & \textbf{91.5} \\ \hline

        \multirow{3}{*}{\Zh} & mBERT & 23.2 & 49.0 & 56.9 & 20.9 & 46.5 & 54.0 & 12.3 & 29.5 & 36.1 & 28.3 & 43.8 & 49.3 \\ 

        & XLM-R & \textbf{30.9} & 57.0 & 63.4 & \textbf{28.8} & 54.6 & 60.9 & 13.9 & 31.9 & 37.9 & 32.5 & 51.1 & 55.9 \\ 
        & LASER2/3 & 10.5 & 19.6 & 24.0 & 10.9 & 21.2 & 24.9 & 7.1 & 16.5 & 22.5 & 7.9 & 16.9 & 22.0  \\
        & LaBSE & 27.1 & \textbf{64.8} & \textbf{73.7} & \textbf{28.8} & \textbf{66.4} & \textbf{75.4} & \textbf{42.1} & \textbf{63.4} & \textbf{69.7} & \textbf{41.0} & \textbf{64.0} & \textbf{71.1} \\ \hline

        \multirow{3}{*}{\Tr} & mBERT & 34.5 & 52.6 & 60.4 & 24.6 & 42.3 & 51.5 & 47.0 & 58.1 & 62.5 & 37.0 & 56.9 & 63.2 \\

        & XLM-R & 35.9 & 62.8 & 69.0 & 28.2 & 50.9 & 59.2 & 50.9 & 62.8 & 66.0 & 45.8 & 60.3 & 64.2  \\
        & LASER2/3 & 35.3 & 57.9 & 64.3 & 32.5 & 50.6 & 57.8 & 56.4 & 67.8 & 70.8 & 45.4 & 65.6 & 70.2 \\
        & LaBSE & \textbf{36.5} & \textbf{71.1} & \textbf{78.5} & \textbf{36.3} & \textbf{74.0} & \textbf{79.9} & \textbf{64.0} & \textbf{80.9} & \textbf{84.3} & \textbf{62.4} & \textbf{80.3} & \textbf{83.6}  \\ \hline
    \end{tabular}
    }

\end{table*}

\newcommand{\dgc}{0.8}
\newcommand{\lgc}{0.9}
\definecolor{dg}{rgb}{\dgc, \dgc, \dgc}
\definecolor{lg}{rgb}{\lgc, \lgc, \lgc}
\newcommand{\Jou}{\cellcolor[gray]{\dgc}}
\newcommand{\Wic}{\cellcolor[gray]{\lgc}}
\newcommand{\capBox}[2]{%
  \begingroup\setlength{\fboxsep}{0.8pt}%
  \colorbox{#2}{\texttt{\hspace*{0.5pt}\vphantom{Ay}\footnotesize #1\hspace*{0.5pt}}}%
  \endgroup
}

\begin{table*}[!ht] 
     \caption{Top-1 BLI Retrieval accuracy for different Retrieval Criteria. 
    Results in cells shaded in \capBox{\strut dark gray}{dg} were obtained from \citet{joulin-etal-2018-loss} while the results in cells shaded in \capBox{\strut light gray}{lg} were obtained from \citet{wickramasinghe-de-silva-2023-sinhala}. All other results in the table are from our own experiments.}
    \label{retrieval_methods_comparison}
    
    \centering
    \resizebox{1\columnwidth}{!}{%
    \begin{tabular}{l|cc|cc|cc|cc|cc|cc|cc|cc|cc|cc}
    \hline
       \multirow{2}{*}{Method} & \multicolumn{20}{c}{language pairs} \\
        \hhline{~--------------------}

        & en-es & es-en & en-fr & fr-en & en-de & de-en & en-ru & ru-en & en-it & it-en & en-si & si-en & en-zh & zh-en & en-ta & ta-en & en-ja & ja-en & en-tr & tr-en \\
        \hline

        Adv.+refine+NN & \Jou 79.1 & \Jou \Jou 78.1 & \Jou 78.1 & \Jou 78.2 & \Jou 71.3 & \Jou 69.6 & \Jou 37.3 & \Jou 45.3 & - & - & - & - & \Jou 30.9 & \Jou 21.9 & - & - & - & - & - & - \\

        Adv.+refine+CSLS & \Jou 81.7 & \Jou 83.3 & \Jou 82.3 & \Jou 82.1 & \Jou 74.0 & \Jou 72.2 & \Jou 44.0 & \Jou 59.1 & - & - & - & - & \Jou 32.5 & \Jou 31.4 & - & - & - & -& - & - \\ \hline
        
        Procrustes+NN & \Jou 77.4 & \Jou 77.3 & \Jou 74.9 & \Jou 76.1 & \Jou 68.4 & \Jou 67.7 & \Jou 47.0 & \Jou 58.2 & 73.0 & 73.6 & \Wic 16.4 & \Wic 21.3 & \Jou 40.6 & \Jou 30.2 & 14.7 & 20.5 & 46.9 & 31.4 & 41.8 & 52.9 \\
        
        Procrustes+CSLS & \Jou 81.4 & \Jou 82.9 & \Jou 81.1 & \Jou 82.4 & \Jou 73.5 & \Jou 72.4 & \Jou 51.7 & \Jou 63.7 & 76.5 & 77.5 & \Wic 20.4 & \Wic 18.0 & \Jou 42.7 & \Jou 36.7 & 16.7 & 22.4 & 52.6 & 38.5 & 47.3 & 59.7 \\ \hline
        
        RCSLS+NN & \Jou 81.1 & \Jou 84.9 & \Jou 80.5 & \Jou 80.5 & \Jou 75.0 & \Jou 72.3 & \Jou 55.3 & \Jou 67.1 & 75.5 & 78.7 & \Wic 21.5 & \Wic 23.3 & \Jou 43.6 & \Jou 40.1 & 17.1 & 23.3 & 22.6 & 0.1 & 46.9 & 59.1 \\
        
        RCSLS+CSLS & \Jou 84.1 & \Jou 86.3 & \Jou 83.3 & \Jou 84.1 & \Jou 79.1 & \Jou 76.3 & \Jou \textbf{57.9} & \Jou 67.2 & 78.3 & 80.3 & \Wic 22.6 & \Wic 19.4 & \Jou 45.9 & \Jou 46.4 & 19.3 & 23.2 & 7.9 & 0.1 & 52.1 & 61.7 \\
        \hline
        \hline

        VecMap+NN & 79.5 & 84.8 & 79.6 & 81.9 & 72.1 & 74.9 & 50.4 & 68.0 & 76.1 & 80.8 & 13.2 & \textbf{45.7} & 39.6 & 43.3 & 17.5 & 33.2 & 48.1 & 41.4 & 46.7 & 63.9 \\
        VecMap+CSLS & 81.3 & 86.5 & 81.9 & 85.3 & 74.5 & 76.3 & 52.7 & 72.1 & 78.8 & 83.3 & 18.1 & 43.2 & 43.3 & 49.6 & 20.2 & 34.7 & \textbf{52.8} & 46.0 & 51.7 & 69.2 \\
        \hline
        
        C1+NN & 81.6 & 84.4 & 81.3 & 82.1 & 76.3 & 74.5 & 56.1 & 67.1 & 77.0 & 81.0 & 17.7 & 17.1 & 41.5 & 44.3 & 20.6 & 26.3 & 30.0 & 34.0 & 52.1 & 64.8 \\
        C1+CSLS &82.1 & 86.1 & 82.3 & 84.4 & 76.5 & 76.5 & 55.4 & 70.6 & 78.6 & 82.3 & 20.7 & 23.3 & 47.8 & 48.2 & 23.1 & 29.7 & 40.8 & 41.4 & 56.5 & 67.6  \\
        \hline

        LaBSE+NN & 40.0 & 65.3 & 48.7 & 72.4 & 45.3 & 56.0 & 17.4 & 48.3 & 41.5 & 64.5 & 4.5 & 35.9 & 19.8 & 42.1 & 15.7 & 35.7 & 6.4 & 26.7 & 36.5 & 64.0 \\

        LaBSE+CSLS & 40.5 & 65.0 & 49.2 & 71.3 & 45.7 & 55.1 & 17.7 & 48.9 & 41.8 & 64.0 & 5.7 & 27.1 & 19.9 & 41.0 & 15.9 & 36.1 & 6.9 & 25.6 & 36.3 & 62.4 \\ \hline

        C2+LaBSE+NN & \textbf{84.9} & 88.1 & \textbf{85.5} & 88.2 & \textbf{80.4} & \textbf{82.1} & 56.1 & 66.0 & \textbf{83.1} & 87.2 & 21.6 & 17.6 & 54.9 & 56.6 & 25.4 & 36.0 & 42.3 & 43.4 & 62.7 & 76.5 \\
        C2+LaBSE+CSLS & 83.3 & \textbf{89.6} & 83.8 & \textbf{89.0} & 77.7 & 81.9 & 55.0 & \textbf{70.7} & 81.3 & \textbf{88.1} & \textbf{25.4} & 25.9 & \textbf{56.6} & \textbf{60.9} & \textbf{30.1} & \textbf{43.5} & 47.3 & \textbf{52.8} & \textbf{65.1} & \textbf{79.3} \\
        \hline
    \end{tabular}
    }

\end{table*}

Table~\ref{retrieval_methods_comparison} compares different alignment techniques and different retrieval criteria. It contains the top-1 retrieval BLI accuracy for all the language pairs. C2+LaBSE-aligned embeddings show the best scores for most of the language pairs. The \emph{cross-domain similarity local scaling} \citet{conneau2017word} (CSLS) criterion outperforms the \emph{nearest neighbor} (NN) criterion in most cases for all the alignment techniques other than C2+LaBSE, while for C2+LaBSE both NN and CSLS show competitive results. In the baseline study, \citet{li-etal-2022-improving} have used the $[CLS]$ token embeddings of mBERT, XLM, and mT5 as the contextualized multilingual embeddings but as per Table~\ref{different_multilingual_models_comparison} we directly used LaBSE output embeddings for the C2 alignment.


\begin{table*}[!htb]
\caption{The Effect of Vocabulary Pruning for BLI score. Some numbers may not be identical to the numbers in Table~\ref{retrieval_methods_comparison} due to minor mismatches in reproducing setups. Also in Table~\ref{retrieval_methods_comparison} \citet{wickramasinghe-de-silva-2023-sinhala} have used cc-FastText vocabulary for the En-Si pair but here we have used wiki-FastText vocabularies for all languages if not otherwise specified}
    \label{tab:Effect_of_Vocabulary_Pruning}
    \centering
    \resizebox{1\columnwidth}{!}{%
    \begin{tabular}{l|cc|cc|cc|cc|cc}
    \hline
    Method & \makecell{RCSLS\\+NN} & \makecell{RCSLS\\+CSLS} & \makecell{LaBSE\\+NN} & \makecell{LaBSE\\+CSLS} & \makecell{VecMap\\+NN} & \makecell{VecMap\\+CSLS} & \makecell{C1\\+NN} & \makecell{C1\\+CSLS} & \makecell{C2\\+NN} & \makecell{C2\\+CSLS} \\ \hline

    En-Ru & 55.4 & 57.1 & 17.4 & 17.7 & \textbf{50.4} & \textbf{52.7} & 56.1 & 55.4 & 56.1 & 55.0 \\
    En-Ru (Pruned) & \textbf{55.5} & \textbf{57.5} & \textbf{43.8} & \textbf{43.7} & 50.1 & 52.6 & \textbf{55.9} & \textbf{55.5} & \textbf{56.8} & \textbf{55.7} \\ \hline
    
    En-Si & \textbf{15.3} & \textbf{17.5} & 4.5 & 5.7 & \textbf{13.2} & \textbf{18.1} & \textbf{17.7} & \textbf{20.7} & 21.6 & 25.4 \\
    En-Si (Pruned) & \textbf{15.3} & \textbf{17.5} & \textbf{46.6} & \textbf{47.7} & \textbf{13.2} & \textbf{18.1} & \textbf{17.7} & \textbf{20.7} & \textbf{21.9} & \textbf{26.1} \\ \hline
    
    En-Zh & 43.3 & 35.4 & 19.8 & 19.9 & 39.6 & 43.3 & 41.5 & 47.8 & 54.9 & 56.6 \\
    En-Zh (Pruned) & \textbf{48.4} & \textbf{40.2} & \textbf{27.1} & \textbf{28.8} & \textbf{42.7} & \textbf{46.3} & \textbf{47.3} & \textbf{51.7} & \textbf{59.4} & \textbf{59.3} \\ \hline
    
    En-Ta & 15.5 & 17.4 & 15.7 & 15.9 & 17.5 & 20.2 & 20.6 & 23.1 & 25.4 & 30.1 \\
    En-Ta (Pruned) & \textbf{17.1} & \textbf{19.3} & \textbf{36.2} & \textbf{36.9} & \textbf{19.3} & \textbf{22.5} & \textbf{22.8} & \textbf{25.3} & \textbf{27.2} & \textbf{31.7} \\ \hline
    
    En-Ja & 6.1 & 4.9 & 6.4 & 6.9 & 48.1 & 52.8 & 30.0 & 40.8 & 42.3 & 47.3 \\
    En-Ja (Pruned) & \textbf{22.6} & \textbf{7.9} & \textbf{37.8} & \textbf{40.1} & \textbf{48.7} & \textbf{55.4} & \textbf{30.9} & \textbf{43.4} & \textbf{45.1} & \textbf{54.2} \\ \hline
    \end{tabular}
    }

\end{table*}

\begin{figure}[!htb]
\centering
\includegraphics[width=1\columnwidth,trim={0 0 0 1.5cm},clip]{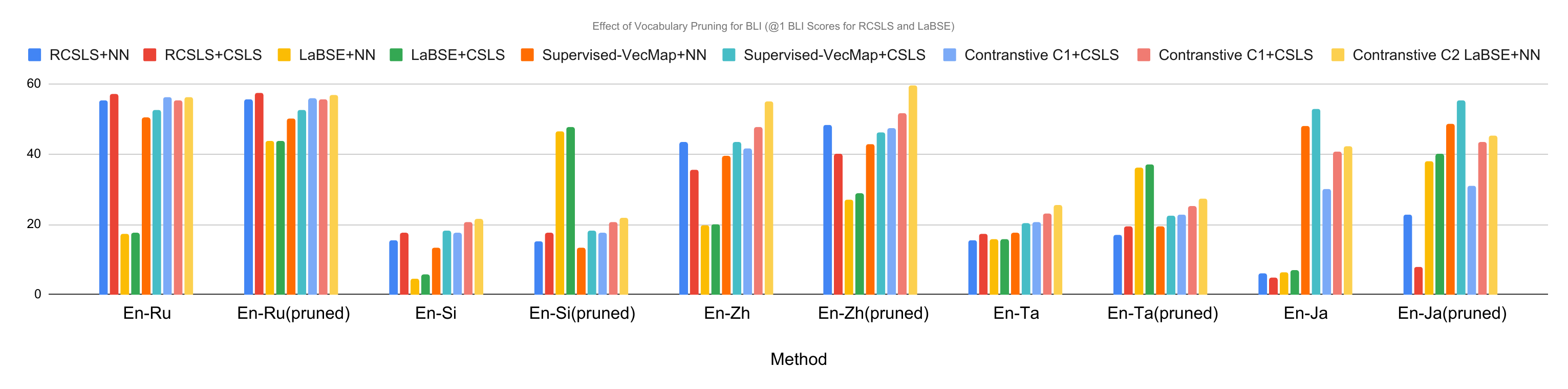}
\caption{Effect of Vocabulary Pruning for Retrieval Accuracy.}
\label{fig:Effect_of_Vocabulary_Pruning}
\end{figure}
        
Figure~\ref{fig:Effect_of_Vocabulary_Pruning} and Table~\ref{tab:Effect_of_Vocabulary_Pruning} show how vocabulary pruning can benefit the BLI score and give a proper idea about the alignment, as we explained in section~\ref{pruning_method}. The experiment has been carried out using all the alignment techniques we have evaluated here. In essence, what we try to study with this experiment is how impurities in the vocabularies could affect BLI. We evaluate BLI with and without vocabulary impurities (i.e., words of other languages than the two languages we evaluate) and check how much the BLI score can be degraded when impurities are there. A significant improvement can be seen in LaBSE contextualized multilingual embeddings when those impurities are removed (see Table~\ref{tab:Improvements_of_Vocabulary_Pruning} for improved percentages).

\begin{table*}[!htb]
    \caption{Impact of the inflections of Sinhala language for alignment evaluation. \emph{Soft} means the stem-based evaluation explained in Section~\ref{inflected_lang_evaluation}. The soft matching is done in the $En \rightarrow Si$ direction only since $\En$ is not a highly inflected language~\cite{mauvcec2004modelling}.}
    \label{impact_of_inflection}
    \centering
    \resizebox{1\columnwidth}{!}{%
    \begin{tabular}{l|ccc|ccc|ccc|ccc}
    \hline
        \multirow{3}{*}{Method} & \multicolumn{6}{c|}{wiki} & \multicolumn{6}{c}{cc} \\
        \hhline{~------------}
        & \multicolumn{3}{c|}{En-Si} & \multicolumn{3}{c|}{Si-En} & \multicolumn{3}{c|}{En-Si} & \multicolumn{3}{c}{Si-En} \\
        \hhline{~------------}
         & P@1 & P@5 & P@10 & P@1 & P@5 & P@10 & P@1 & P@5 & P@10 & P@1 & P@5 & P@10 \\  \hline

        RCSLS + NN & 15.3 & 30.4 & 37.5 & 13.2 & 34.1 & 43.3 & 21.5 & 40.9 & 48.3 & 23.3 & 44.9 & 53.2 \\ 

        RCSLS + NN (Soft) & 21.3 & 36.7 & 39.6 & 13.2 & 34.1 & 43.3 & 30.7 & 51.1 & 58.7 & 23.3 & 44.9 & 53.2  \\ \hline
        
        RCSLS + CSLS & 17.5 & 33.4 & 41.3 & 15.5 & 29.3 & 35.9 & 22.6 & 42.3 & 49.1 & 19.4 & 35.4 & 42.1 \\ 

        RCSLS + CSLS (Soft) & 23.4 & 43.7 & 47.1 & 15.5 & 29.3 & 35.9 & 32.3 & 51.7 & 59.20 & 19.4 & 35.4 & 42.1  \\ \hline
        
        LaBSE + NN & 46.6 & 70.9 & 77.4 & \textbf{35.9} & \textbf{64.4} & \textbf{73.9} & 36.3 & 63.9 & 70.9 & \textbf{41.0} & \textbf{69.6} & \textbf{75.5} \\ 

        LaBSE + NN (Soft) & 57.8 & 77.1 & 81.7 & \textbf{35.9} & \textbf{64.4} & \textbf{73.9} & \textbf{52.4} & \textbf{74.3} & 79.0 & \textbf{41.0} & \textbf{69.6} & \textbf{75.5} \\ \hline

        LaBSE + CSLS & 47.7 & 72.7 & 78.9 & 27.1 & 55.8 & 69.2 & 35.0 & 62.9 & 71.0 & 31.8 & 62.3 & 71.3 \\
        LaBSE + CSLS (Soft) & \textbf{58.9} & \textbf{79.2} & \textbf{83.5} & 27.1 & 55.8 & 69.2 & 50.7 & 73.8 & \textbf{79.9} & 31.8 & 62.3 & 71.3 \\ \hline
        
    \end{tabular}
    }

\end{table*}

\begin{figure}[!ht]
\centering
\includegraphics[width=1\columnwidth,trim={0 0 0 1.5cm},clip]{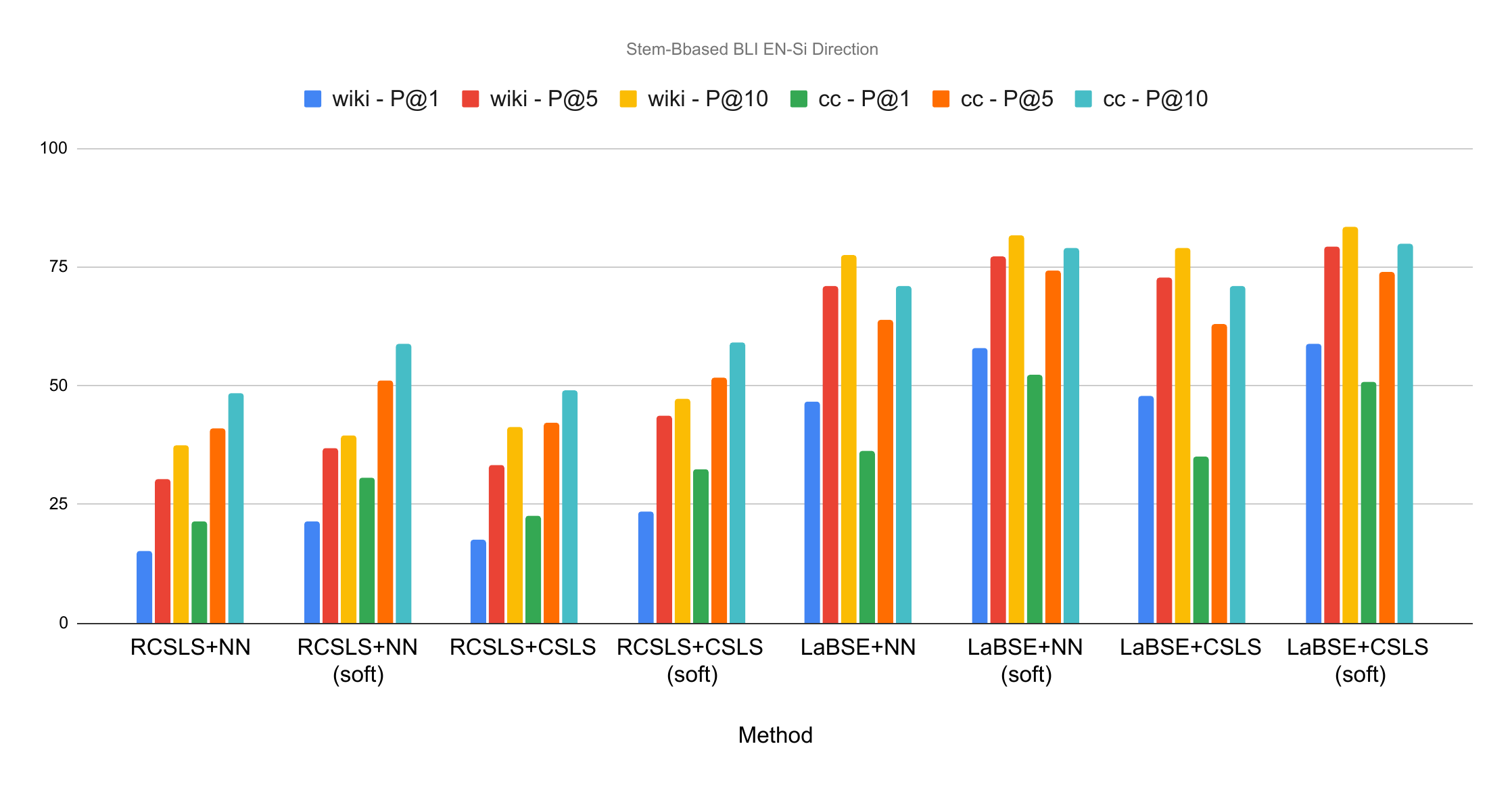}
\caption{Impact of Inflection for BLI using Sinhala(\Si)-English(\En) pair.\\\emph{soft}: our proposed stem-based BLI. We have shown the results for En$\rightarrow$Si direction. \emph{Wiki}: Wikipedia FastText vocabulary, and \emph{cc}: common-crawl FastText vocabulary.}
\label{fig:Effect_of_inflection_for_BLI}
\end{figure}

\begin{table}[!ht]
\caption{LaBSE Embedding-based BLI Score Improvements after Vocabulary Pruning}
    \label{tab:Improvements_of_Vocabulary_Pruning}
    \centering
        \begin{tabular}{l|rr}
        \hline
        Language & LaBSE+NN & LaBSE+CSLS \\ \hline
        En-Ru & 150\% & 145\% \\
        En-Si & 935\% & 736\% \\
        En-Zh & 36\% & 45\% \\
        En-Ta & 130\% & 132\% \\
        En-Ja & 490\% & 480\% \\ \hline
        \end{tabular}

\end{table}

\begin{table}[!ht]
\caption{En-Si BLI Score Improvements by Stem-based BLI}
    \label{tab:Improvements_of_Stem_BLI}
    \centering
        \begin{tabular}{l|rrr|rrr}
        \hline
        \multirow{2}{*}{Method} & \multicolumn{3}{c|}{wiki En-Si} & \multicolumn{3}{c}{cc En-Si} \\
        \hhline{~------}
        & P@1 & P@5 & P@10 & P@1 & P@5 & P@10 \\
        \hline
        RCSLS+NN & 39\% & 21\% & 6\% & 43\% & 25\% & 22\% \\
        RCSLS+CSLS & 34\% & 31\% & 14\% & 43\% & 22\% & 21\% \\
        LaBSE+NN & 24\% & 9\% & 6\% & 44\% & 16\% & 11\% \\
        LaBSE+CSLS & 23\% & 9\% & 6\% & 45\% & 17\% & 13\% \\ \hline
        \end{tabular}

\end{table}

Figure~\ref{fig:Effect_of_inflection_for_BLI} shows the effectiveness of our proposed \emph{Stem-based BLI} technique which we explained in Section~\ref{inflected_lang_evaluation}. We have used Sinhala as the candidate language to evaluate this. Sinhala is a highly inflected language which is closely related to one of the highly inflected languages, Sanskrit\footURL{https://en.wikipedia.org/wiki/Sanskrit}~\cite{de2019survey}. The stem-based matching is valid only in the $\En \rightarrow \Si$ direction since English is not a highly inflected language~\cite{mauvcec2004modelling}. Also, we have carried out the experiment using only RCSLS and LaBSE techniques but, we can see that there is a clear score improvement when stem-based matching is used. This result improvement has been shown using both Wikipedia (wiki) and Common-Crawl (cc) FastText models. These results show that the standard BLI task can output pessimistic results in inflected language cases and the proposed stem-based BLI is more realistic in such cases (see Table~\ref{impact_of_inflection} and Table~\ref{tab:Improvements_of_Stem_BLI}).

Obtaining the best BLI scores for C2 can be expected because we are trying to get the best out of either of the embedding types using C2. It is a two-stage procedure in which we can start with any type of aligned static embeddings and then we can further improve the alignment using any type of multilingual contextualized embeddings. Here in our study, we have used VecMap as the starting embeddings and LaBSE as the contextualized embeddings for C2.

\section{Discussion}

Our main objective in this study was to evaluate how good the widely used BLI metric is as a measurement for the degree of alignment between two embedding spaces.
To that end, we measured the standard BLI score for aligned monolingual word embeddings, multilingual embeddings, and combined embeddings. We brought all three types of embeddings onto a common comparable stage by narrowing down the study to the word embedding paradigm (contextualized models give fixed embeddings when a single word is fed and therefore they behave akin to static embedding models), selecting the same vocabulary for the evaluation, and using the same evaluation datasets. Accordingly, we have compared RCSLS, VecMap, C1, LaBSE, and C2+LaBSE aligned embeddings. From the results thereof, we observe several behaviours and draw several conclusions. 

According to Table~\ref{retrieval_methods_comparison}, we observed that in most of the cases, C2+LaBSE alignment which is a combined alignment technique of aligned static embeddings and multilingual contextualized embeddings has given the best results. RCSLS, VecMap, and C1 traditional aligned monolingual embeddings have shown slightly low but equally competitive BLI scores in most of the cases 
while especially in Sinhala and Tamil cases (which are two low-resource and inflected languages), LaBSE (multilingual embeddings) has shown somewhat competitive results with traditional alignment techniques.
Therefore from the standard BLI measurements what we can observe is that the quality of alignment of multilingual embeddings lags behind traditional and combined embeddings. According to the standard BLI scores, when it comes to inflected languages such as Sinhala and Tamil, it demonstrates very poor levels of alignment compared to other language pairs.

One of the reasons for the lag of contextualized multilingual embedding BLI scores could be that while trying to achieve a collective alignment for multiple languages, the alignment of a given two language pairs can be compromised when training. Also, the high-resource language focus is compromised by the data sampling techniques followed to focus on low-resource languages. (see Section~\ref{mono_vs_multi}). 

When vocabulary pruning is applied, we see a huge improvement in the contextualized embedding technique (36\% - 935\% see Table~\ref{tab:Improvements_of_Vocabulary_Pruning}). The reason for this can be postulated as follows; since the multilingual embeddings are already aligned for all the languages it supports, for a given source word, the closest word can be from any of these supported languages. Therefore in order to remove the impact of redundant languages and only to focus on the language pair of interest we can use the proposed vocabulary pruning. 
We can claim that the actual degree of alignment of multilingual models cannot be properly measured using BLI if all the languages are considered in the experiment. Declaring $@N$ on a multilingual model of $N$ languages to be equivalent to $@1$ of a monolingual model may mathematically guarantee a hit at the optimal embedding alignment scenario, but this would need a different degree of experiments that is out of scope for this study and is thus kept as future work.  

We showed that standard BLI is not as informative for high-inflected languages as much as it is for low-inflected languages, using the Sinhala language as an example of the former.
Thus, for highly inflected languages, it can be claimed that it is better to go with a soft matching evaluation technique to evaluate the alignment quality since attempting to find exact matches will under-report the true quality of aligned models. Interestingly, in our study, we do not identify any clear impact of the language family on the alignment results.


\bibliographystyle{splncs04nat}
\bibliography{iccci_custom}
%
%
%
%




\end{document}